\pdfoutput=1
\documentclass[11pt]{article}
\usepackage[final]{acl}

\usepackage{times}
\usepackage{latexsym}

\usepackage[T1]{fontenc}
\usepackage[utf8]{inputenc}

\usepackage{microtype}

\usepackage{inconsolata}

\usepackage{graphicx}
\usepackage{subcaption}

\usepackage{listings}
\usepackage{xcolor}
\usepackage{pgffor} % if you keep the foreach loop

\usepackage{tcolorbox}
\usepackage{enumitem}
\usepackage{booktabs}
\usepackage{multirow}
\title{ITUNLP at SemEval-2025 Task 8: Question-Answering over Tabular Data: A Zero-Shot Approach using LLM-Driven Code Generation}

\author{
    Atakan Site\thanks{These authors contributed equally.}\thanks{Corresponding author.}, 
    Emre Hakan Erdemir$^{*}$, 
    Gülşen Eryiğit \\
    Department of Artificial Intelligence and Data Engineering \\ 
    Istanbul Technical University \\
    \texttt{\{site21, erdemire21, gulsenc\}@itu.edu.tr}
}

\begin{document}
\maketitle
\begin{abstract}
This paper presents our system for SemEval-2025 Task 8: DataBench, Question-Answering over Tabular Data. The primary objective of this task is to perform question answering on given tabular datasets from diverse domains under two subtasks: DataBench QA (Subtask I) and DataBench Lite QA (Subtask II). %Models must answer the given questions according to the expected answer formats. The task comprises two subtasks: DataBench QA (Subtask I), where models must answer questions using the full dataset, and DataBench Lite QA (Subtask II), which limits responses to a sampled subset of at most 20 rows. 
To tackle both subtasks, we developed a zero-shot solution with a particular emphasis on leveraging Large Language Model (LLM)-based code generation. Specifically, we propose a Python code generation framework utilizing state-of-the-art open-source LLMs to generate executable Pandas code via optimized prompting strategies. Our experiments reveal that different LLMs exhibit varying levels of effectiveness in Python code generation. Additionally, results show that Python code generation achieves superior performance in tabular question answering compared to alternative approaches. %Although our ranking within zero-shot systems is not known at the time of this paper's submission, our system, which addresses both subtasks, achieved eighth place in Subtask I and sixth place in Subtask II within the open-source models category, ranking among the 30 systems that outperformed the baseline.
Although our ranking among zero-shot systems is unknown at the time of this paper's submission, our system achieved eighth place in Subtask I and sixth place in Subtask~II among the 30 systems that outperformed the baseline in the open-source models category.

\end{abstract}

\section{Introduction}

Question Answering (QA) is a fundamental task in Natural Language Processing (NLP), where the most relevant answers are retrieved from a given document or plain text. Apart from such unstructured data, working with widely used structured data is crucial for real-world applications. Moreover, structured data encompasses a much broader semantic scope. One important form of structured data is tabular data, which consists of rows with a consistent set of features. Unlike unstructured documents, tabular data exhibits complex and heterogeneous relationships that require specialized processing techniques. Information retrieval from tabular data is typically performed using various SQL queries and similar approaches. However, these methods depend on rigid rule-based systems and fail to consider the semantic properties of the data. Consequently, natural-language queries over tabular data face significant limitations. As a result, question-answering systems developed for tabular data have garnered significant interest among researchers.

The process of converting a natural language query into a machine-executable logical form is known as semantic parsing \citep{wang-etal-2015}. Early studies primarily focused on datasets that required adapting specific logical forms for each table structure type. This approach, however, led to suboptimal performance, particularly in tabular structures spanning multiple domains \citep{pasupat2015compositionalsemanticparsingsemistructured}. On the other hand, end-to-end trained transformers are widely employed, as they handle both question/query interpretation and reasoning over tabular data \citep{deng2020turltableunderstandingrepresentation}. The recent advancements in LLMs have become a pivotal focus in tabular question answering, as in many other problem domains. However, LLM-based approaches introduce several challenges, including high computational costs and limited context length, making scalable and efficient tabular QA systems an open research problem. To address these challenges and foster the development of effective tabular question-answering methods, SemEval-2025 Task 8 \citep{osesgrijalba-etal-2025-semeval-2025} has been designed to introduce the necessary level of difficulty through two distinct subtasks.

In this paper, we propose a zero-shot system to address these tasks, focusing primarily on LLM-based code generation. Our approach introduces a unified framework leveraging state-of-the-art open-source LLMs, including DeepSeek-R1 \citep{deepseekai2025deepseekr1incentivizingreasoningcapability}, DeepSeek-V3 \citep{deepseekai2025deepseekv3technicalreport}, Qwen2.5-Coder-32B-Instruct \citep{hui2024qwen25codertechnicalreport}, and Llama-3.3-70B-Instruct \citep{grattafiori2024llama3herdmodels}. We employ efficient prompting strategies to generate executable Python Pandas\footnote{\url{https://pandas.pydata.org/}} library code. To enhance LLM understanding of tabular structures, the generated Python code is executed in a controlled environment. A key feature of our system is its iterative error-handling mechanism. If the initial code execution fails, the error message and faulty code are sent back to the LLM for correction, with a maximum of two iterations. This mechanism significantly improves robustness, reducing failure rates in complex queries.

We observe that one model in our pipeline achieves the highest accuracy on Subtask I (84.67\%), while another leads Subtask II (85.05\%), both without task-specific fine-tuning. %Our best-performing system outperforms the official baseline and other models in our pipeline, achieving 84.67\% accuracy on the test set of Subtask I and 85.05\% accuracy on the test set of Subtask II, without requiring task-specific fine-tuning.
All code is available on our GitHub repository\footnote{\url{https://github.com/erdemire21/semeval8-itunlp}}.

\begin{figure*}[t]
    \centering
    \includegraphics[width=1\linewidth]{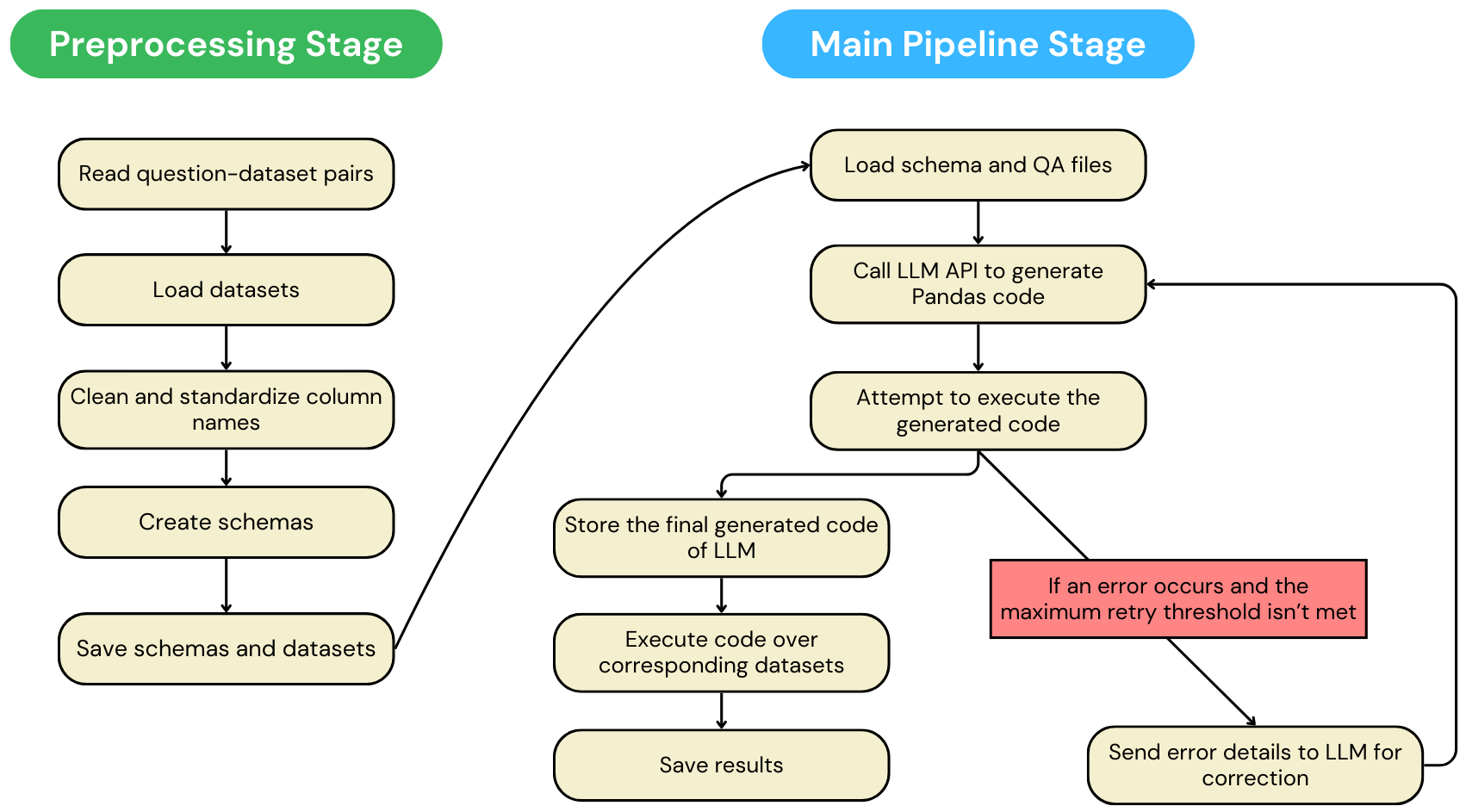}
    \caption{Our proposed framework.}
    \label{fig:system-arch}
\end{figure*}

\section{Related Work}

This section reviews recent developments in LLMs, focusing on their applications in tabular question answering.

In recent years, the emergence of the Transformer architecture \citep{vaswani2017attentionneed} has led to remarkable advancements in language modeling tasks. This progress has resulted in state-of-the-art performance across various NLP tasks. Consequently, the application of transformer architectures to problems requiring tabular modeling has become inevitable. Early studies primarily focused on different embedding mechanisms \citep{yin2020tabertpretrainingjointunderstanding}, pre-training strategies \citep{Wang_2021}, and architectural modifications \citep{huang2020tabtransformertabulardatamodeling}. The core approach introduced by these methods was pre-training Transformer architectures from scratch for tabular data \citep{Herzig_2020}. However, this approach faces efficiency and scalability limitations, particularly when models need to generalize across multiple domains. Generally, pre-trained language models struggle to adapt efficiently to task-specific tabular datasets.

More recently, the emergence of LLMs has brought about a significant transformation in the field. Models such as GPT-3 \citep{brown2020languagemodelsfewshotlearners} and LLaMa \citep{touvron2023llamaopenefficientfoundation} have demonstrated strong few-shot and zero-shot capabilities, achieving state-of-the-art performance across various tasks while often requiring little to no task-specific data. These advancements have enabled the use of a single, unified model for solving complex tabular tasks. The transition from training models from scratch or adapting pre-trained language models to leveraging LLMs represents a significant paradigm shift in tabular data processing. However, the application of LLMs to tabular question answering introduces several challenges. One major limitation is the context length constraint inherent to LLMs. When processing large or multiple tables, the limited context size prevents the model from encoding all necessary information. Additionally, handling multiple tables often leads to hallucinations, where models generate inaccurate or misleading responses.

To overcome these limitations, researchers have leveraged the in-context learning capabilities of LLMs. The effectiveness of LLM-based approaches largely depends on how tabular data and question queries are represented and utilized. For tabular data, appropriate table schemas and prompting strategies incorporating relevant examples are designed to enhance model comprehension. Query representation can also significantly impact performance. A common strategy involves decomposing complex queries into step-by-step subqueries, improving model interpretability \citep{yang-etal-2024-effective}. Another approach is transforming queries into intermediate representations such as Python code or SQL queries, enabling structured execution \citep{cao-etal-2023,zhang-etal-2024-syntqa}. These advancements have led to models capable of performing task-specific reasoning without requiring additional fine-tuning. 

Building on insights from previous studies, we find that effectively addressing SemEval-2025 Task~8 requires a deep understanding of query semantics and table structures, as well as the ability to generate accurate answers across diverse answer formats. Motivated by these challenges, we introduce a novel framework that integrates schema-guided prompting, controlled execution, and an error-handling mechanism. Our extensive evaluations and prompt strategy experiments highlight the effectiveness of our approach in enhancing accuracy and robustness. These findings show the practicality and applicability of the proposed approach in real-world scenarios, where tabular data must be processed dynamically without requiring task-specific fine-tuning. 

\section{Data}
The original DataBench dataset \citep{oses-grijalba-etal-2024-question} provides 1308 questions from 65 different domains, each containing question-answer pairs written in English. During the competition, this dataset was expanded with the addition of a new test split \citep{osesgrijalba-etal-2025-semeval-2025}. The exact dataset statistics are presented in Table \ref{tab:databench_stats}. The train and development splits contain the following columns: 
\setlist{nolistsep}
\begin{itemize}[noitemsep]
    \item \textbf{question}: The natural language question.
    \item \textbf{answer}: The response to the question for DataBench QA subtask.
    \item \textbf{type}: The type of the answer, which can be \textbf{boolean, number, category, list[category], list[number]}.
    \item \textbf{columns\_used}: The columns of the dataset required to answer the question.
    \item \textbf{column\_types}: The data types of these columns, which include {boolean, number (e.g., UInt8, uint32, uint16)}.
    \item \textbf{sample\_answer}: The response to the question for DataBench Lite QA subtask.
    \item \textbf{dataset}: The name of the dataset from which the question is derived.
\end{itemize}

\begin{table}[htbp]
    \centering
    \begin{tabular}{lcc}
        \hline
        \textbf{Split} & \textbf{Questions} & \textbf{Datasets} \\
        \hline
        Train      & 988 & 49 \\
        Dev        & 320 & 16 \\
        Test       & 522 & 15 \\
        \hline
    \end{tabular}
    \caption{DataBench dataset statistics.}
    \label{tab:databench_stats}
\end{table}

The sample\_answer column is specifically included for the DataBench QA Lite subtask, which is a simplified version of the DataBench QA task. This subset consists of 20 sampled entries from the original dataset, serving as a small-scale reference for evaluation.

In contrast to these extensively annotated train and development splits, the test split only has question and dataset columns to ensure proper evaluation without data leak for the competition.

Although the dataset provides structured train and development splits with detailed annotations, this study did not utilize these data for training, as we preferred a zero-shot approach that does not involve fine-tuning.

\section{System Overview}
Our approach involves two main steps in providing an answer to questions over tabular data: preprocessing and then code generation and execution. The complete workflow is illustrated in Figure~\ref{fig:system-arch}.

\subsection{Preprocessing}
Our preprocessing steps include obtaining the given questions and datasets from the competition website, followed by a series of normalization and standardization techniques, and finally creating a schema for each dataset for LLM prompting. Each dataset is transformed with transformation rules. First, all spaces and non-word characters are replaced with underscores except for trailing special characters, which are removed. Second, all column names are converted to lowercase, and duplicate columns are renamed by appending a number to each duplicate. For example, if there are two cols named "col" and "Col@", the second one becomes "col\_2".

After normalization and standardization, we construct a schema for each dataset to enhance the LLM's understanding of the table structure. The schemas include each dataset's name, each column, each column's data type, 5 unique values from each column, and the total unique values that a column contains. The example values are limited to a hundred characters total to avoid excessive verbosity and potential token overload. Examples of the constructed schemas can be seen in Appendix~\ref{sec:example-schemas}, (e.g., see the schema for the TripAdvisor dataset in Appendix~\ref{sec:TripAdvisor-schema}). We use the full dataset for schema creation for both full and sample datasets.

%full yerine baska bir sey yazilabilir
%Subsection yerine appendixte schema örneği olabilir

\subsection{Code Generation and Execution}
The code generation step is done with a prompt that includes the question, detailed instructions and the corresponding dataset schema. A detailed breakdown of the code generation prompt is provided in Appendix~\ref{sec:codegen}. The generated code is executed in a controlled environment, where dynamic imports are extracted, and the execution output is captured in its original format. In cases where execution fails, an error handling mechanism is triggered. The system captures the error message along with the faulty code and sends it to the LLM for automatic correction. The LLM then generates a revised version of the code. This iterative process is run until the predefined threshold is met. If the provided code is still faulty after the maximum number of attempts, execution is terminated for that query. The execution result from the last successfully executed code is then set as the final answer for the corresponding question.

\section{Experimental Setup}

%Error handlingden once ve sonra kac error var yazilabilir
Our zero-shot framework was tested on the officially released development and test datasets of SemEval 2025 Task 8, covering its two subtasks \citep{osesgrijalba-etal-2025-semeval-2025}. The models used in our system were selected based on their performance in code generation tasks, ensuring their effectiveness in handling structured and semi-structured tabular question answering. Additionally, we opted for a maximum of two iterations based on preliminary experiments, which showed that attempts beyond two iterations rarely produced further improvements. To provide a more comprehensive error analysis, we also conducted additional experiments with three iterations. To evaluate system performance, we used Accuracy, the official evaluation metric of SemEval 2025 Task 8. Furthermore, we analyzed the impact of our iterative error-handling mechanism on execution reliability by measuring error reduction rates across different models. These evaluations provide insights into both models accuracy and execution robustness in tabular question answering.

\begin{table*} %[h!]
  \centering
    \begin{tabular}{l cc cc}  
        \hline
        \textbf{Models} & \multicolumn{2}{c}{\textbf{Subtask I (DataBench)}} & \multicolumn{2}{c}{\textbf{Subtask II (DataBench Lite)}} \\
        \cline{2-5}
                         & \textbf{Dev} & \textbf{Test} & \textbf{Dev} & \textbf{Test} \\
        \hline
        DeepSeek-R1      & \textbf{88.43}  & 84.09  & \textbf{86.56}  & \textbf{85.05}  \\
        DeepSeek-V3 & 82.50  & \textbf{84.67}  & 78.75  & 80.84  \\
        
        Qwen2.5-Coder-32B-Instruct & 87.18  & 83.90  & 85.31  & 81.99  \\
        Llama-3.3-70B-Instruct	 & 86.56  & 83.14  & 82.81  & 81.03  \\
        \hline
    \end{tabular}
  \caption{\label{result-table}
    Results on the DataBench subtasks across all models.
  }
\end{table*}

\section{Results}
\label{sec:bibtex}

The performance of the models is presented in Table~\ref{result-table}. Our results indicate that one of the DeepSeek models (i.e., DeepSeek-R1 and DeepSeek-V3)  outperforms all other models across both subtasks.
We see that DeepSeek-V3 falls behind all the others on the development sets, but performs better specifically on the test set of Subtask I. DeepSeek-R1, which is a subsequent iteration, building upon V3 with enhanced capabilities via reinforcement learning, outperforms Qwen2.5-Coder-32B-Instruct and Llama-3.3-70B-Instruct models on all tasks and datasets, falling behind DeepSeek-V3 by 0.52 percentage points on the Subtask I test set.  

%particularly DataBench Lite QA (Subtask II), where the performance gap is more pronounced. 
%Compared to Qwen2.5-Coder-32B-Instruct, DeepSeek-R1 shows a notable improvement in both development and test datasets. Llama-3.3-70B-Instruct and DeepSeek-V3 perform competitively, but their accuracy is consistently lower. 
Moreover, in the official evaluation within the open-source models category, our best-performing model ranked eighth in Subtask I and sixth in Subtask II, placing among the 30 systems that outperformed the baseline. These results further highlight the effectiveness of our approach in zero-shot tabular question answering. At the time of this paper's submission, due to a lack of information on other solutions, we were unable to evaluate our performance relative to other zero-shot systems in the competition. Through our manual observations, we identified that the test datasets are significantly more challenging. However, we do not believe that every question-answer pair in these datasets can perfectly represent the real-world performance of the models. Nonetheless, the widening performance gap in the more challenging test sets suggests that DeepSeek-R1 may generalize to the problem more effectively, providing evidence of its superior adaptability.

\begin{table}
  \centering
  \resizebox{\columnwidth}{!}{%
  \begin{tabular}{lcc}
    \hline
    \textbf{Models} & \textbf{Dev Set} & \textbf{Test Set} \\
    \hline
    DeepSeek-R1                & 9 $\rightarrow$ 6  & 15 $\rightarrow$ 7  \\
    DeepSeek-V3                & 35 $\rightarrow$ 11& 18 $\rightarrow$ 9  \\
    
    Qwen2.5-Coder-32B-Instruct & 11 $\rightarrow$ 9 & 25 $\rightarrow$ 8  \\
    Llama-3.3-70B-Instruct     & 16 $\rightarrow$ 5 & 16 $\rightarrow$ 10 \\
    \hline
  \end{tabular}%
  }
  \caption{The change in the amount of code execution errors before and after the error fixing loop.}
  \label{tab:error_handling_changes_joined}
\end{table}

In addition, as shown in Table~\ref{tab:error_handling_changes_joined}, our error handling mechanism decreases the number of execution errors by nearly half on average, demonstrating not only its effectiveness but also its necessity for ensuring reliable execution. It should be noted that the initial error rate and the accuracy over both tasks show a strong correlation, with models that achieve higher accuracy also generating less faulty code to begin with. This suggests that better-performing models inherently produce more reliable code, thereby reducing the need for iterative error correction loops and improving overall execution efficiency.

To analyze error patterns and the impact of our correction mechanism in greater detail, we grouped errors into three main categories: Runtime, Degenerate Loop, and Syntax. Notably, the Runtime category includes diverse errors such as KeyError and ValueError, but for simplicity, we report them under a single label. Our findings also indicate that some errors transform into different types across iterations.

We define Degenerate Loop errors as cases where an LLM repeatedly generates identical or nearly identical output sequences, continuing indefinitely until it reaches its maximum token limit.

Table~\ref{tab:error-distribution} presents the distribution of error types across models and iterations. Results show that most initial failures are due to Runtime errors, while Syntax and Loop errors are less frequent but may persist across multiple correction attempts. Specifically, Syntax errors are observed exclusively in DeepSeek-R1 and DeepSeek-V3 models, with no such errors detected for Llama-3.3-70B-Instruct or Qwen2.5-Coder-32B-Instruct across any dataset or iteration.

Similarly, Degenerate Loop errors are observed solely in DeepSeek-R1 and DeepSeek-V3, with no occurrences in Llama-3.3-70B-Instruct or Qwen2.5-Coder-32B-Instruct. As shown in Figures~\ref{fig:system-arch2} and~\ref{fig:system-arch3}, although some Degenerate Loop errors are corrected, a notable portion still results in failures.

Finally, Figure~\ref{fig:system-arch2} provides an overview of error resolution across iterations, showing that most runtime errors are resolved within the first two attempts. Figure~\ref{fig:system-arch3} further breaks down specific error types, such as FileNotFoundError, KeyError, and NameError, offering a more fine-grained view.

%Since the test dataset's answers have not been available to the participants at the time of submission of this paper, we may not provide sufficient analysis to explain the drop in accuracies between SubTask I and SubTask II test sets 

\section{Conclusions}

In conclusion, this paper presented the solution developed by the ITUNLP group for SemEval-2025 Task 8. 
The proposed approach addressed the tabular question answering task in zero-shot scenarios. 
Our method yields promising results in zero-shot tabular question answering, achieving higher ranks ($8^{th}$ place in Subtask I and $6^{th}$ in Subtask~II) within the 30 participant systems in the open-source category.
Since these 30 systems may have employed fine-tuning or few-shot learning techniques,
further analysis would be possible upon the publication of the system description papers that achieved better results on the same category of the shared task, which will  provide a clearer understanding of our ranking within zero-shot frameworks.
%Experimental results demonstrate the superiority of our method, providing a comprehensive comparison with state-of-the-art open-source LLMs. 

As this study focuses only on open-source LLMs, future work could include evaluating proprietary LLMs within our proposed framework to gain a broader perspective on model performance. 
%Additionally, due to the time constraints, we were unable to test our framework on a wider range of datasets. 
Furthermore, the DataBench dataset consists of questions that require using only a single table. As future work, we aim to evaluate our zero-shot model's performance on multi-table reasoning tasks, further expanding its applicability.

\bibliography{anthology,custom}

\begin{thebibliography}{19}
\providecommand{\natexlab}[1]{#1}

\bibitem[{AI@Meta(2024)}]{grattafiori2024llama3herdmodels}
AI@Meta. 2024.
\newblock \href {https://arxiv.org/abs/2407.21783} {The llama 3 herd of models}.
\newblock \emph{Preprint}, arXiv:2407.21783.

\bibitem[{Brown et~al.(2020)Brown, Mann, Ryder, Subbiah, Kaplan, Dhariwal, Neelakantan, Shyam, Sastry, Askell, Agarwal, Herbert-Voss, Krueger, Henighan, Child, Ramesh, Ziegler, Wu, Winter, Hesse, Chen, Sigler, Litwin, Gray, Chess, Clark, Berner, McCandlish, Radford, Sutskever, and Amodei}]{brown2020languagemodelsfewshotlearners}
Tom~B. Brown, Benjamin Mann, Nick Ryder, Melanie Subbiah, Jared Kaplan, Prafulla Dhariwal, Arvind Neelakantan, Pranav Shyam, Girish Sastry, Amanda Askell, Sandhini Agarwal, Ariel Herbert-Voss, Gretchen Krueger, Tom Henighan, Rewon Child, Aditya Ramesh, Daniel~M. Ziegler, Jeffrey Wu, Clemens Winter, Christopher Hesse, Mark Chen, Eric Sigler, Mateusz Litwin, Scott Gray, Benjamin Chess, Jack Clark, Christopher Berner, Sam McCandlish, Alec Radford, Ilya Sutskever, and Dario Amodei. 2020.
\newblock \href {https://arxiv.org/abs/2005.14165} {Language models are few-shot learners}.
\newblock \emph{Preprint}, arXiv:2005.14165.

\bibitem[{Cao et~al.(2023)Cao, Chen, Liu, Wang, and Fried}]{cao-etal-2023}
Yihan Cao, Shuyi Chen, Ryan Liu, Zhiruo Wang, and Daniel Fried. 2023.
\newblock \href {https://doi.org/10.18653/v1/2023.emnlp-main.897} {{API}-assisted code generation for question answering on varied table structures}.
\newblock In \emph{Proceedings of the 2023 Conference on Empirical Methods in Natural Language Processing}, pages 14536--14548, Singapore. Association for Computational Linguistics.

\bibitem[{DeepSeek-AI et~al.(2025{\natexlab{a}})DeepSeek-AI, Guo, Yang, Zhang, Song, Zhang, Xu, Zhu, Ma, Wang, and et~al.}]{deepseekai2025deepseekr1incentivizingreasoningcapability}
DeepSeek-AI, Daya Guo, Dejian Yang, Haowei Zhang, Junxiao Song, Ruoyu Zhang, Runxin Xu, Qihao Zhu, Shirong Ma, Peiyi Wang, and et~al. 2025{\natexlab{a}}.
\newblock \href {https://arxiv.org/abs/2501.12948} {Deepseek-r1: Incentivizing reasoning capability in llms via reinforcement learning}.
\newblock \emph{Preprint}, arXiv:2501.12948.

\bibitem[{DeepSeek-AI et~al.(2025{\natexlab{b}})DeepSeek-AI, Liu, Feng, Xue, Wang, Wu, Lu, Zhao, Deng, Zhang, and et. al.}]{deepseekai2025deepseekv3technicalreport}
DeepSeek-AI, Aixin Liu, Bei Feng, Bing Xue, Bingxuan Wang, Bochao Wu, Chengda Lu, Chenggang Zhao, Chengqi Deng, Chenyu Zhang, and et. al. 2025{\natexlab{b}}.
\newblock \href {https://arxiv.org/abs/2412.19437} {Deepseek-v3 technical report}.
\newblock \emph{Preprint}, arXiv:2412.19437.

\bibitem[{Deng et~al.(2020)Deng, Sun, Lees, Wu, and Yu}]{deng2020turltableunderstandingrepresentation}
Xiang Deng, Huan Sun, Alyssa Lees, You Wu, and Cong Yu. 2020.
\newblock \href {https://arxiv.org/abs/2006.14806} {Turl: Table understanding through representation learning}.
\newblock \emph{Preprint}, arXiv:2006.14806.

\bibitem[{Herzig et~al.(2020)Herzig, Nowak, Müller, Piccinno, and Eisenschlos}]{Herzig_2020}
Jonathan Herzig, Pawel~Krzysztof Nowak, Thomas Müller, Francesco Piccinno, and Julian Eisenschlos. 2020.
\newblock \href {https://doi.org/10.18653/v1/2020.acl-main.398} {Tapas: Weakly supervised table parsing via pre-training}.
\newblock In \emph{Proceedings of the 58th Annual Meeting of the Association for Computational Linguistics}. Association for Computational Linguistics.

\bibitem[{Huang et~al.(2020)Huang, Khetan, Cvitkovic, and Karnin}]{huang2020tabtransformertabulardatamodeling}
Xin Huang, Ashish Khetan, Milan Cvitkovic, and Zohar Karnin. 2020.
\newblock \href {https://arxiv.org/abs/2012.06678} {Tabtransformer: Tabular data modeling using contextual embeddings}.
\newblock \emph{Preprint}, arXiv:2012.06678.

\bibitem[{Hui et~al.(2024)Hui, Yang, Cui, Yang, Liu, Zhang, Liu, Zhang, Yu, Lu, Dang, Fan, Zhang, Yang, Men, Huang, Zheng, Miao, Quan, Feng, Ren, Ren, Zhou, and Lin}]{hui2024qwen25codertechnicalreport}
Binyuan Hui, Jian Yang, Zeyu Cui, Jiaxi Yang, Dayiheng Liu, Lei Zhang, Tianyu Liu, Jiajun Zhang, Bowen Yu, Keming Lu, Kai Dang, Yang Fan, Yichang Zhang, An~Yang, Rui Men, Fei Huang, Bo~Zheng, Yibo Miao, Shanghaoran Quan, Yunlong Feng, Xingzhang Ren, Xuancheng Ren, Jingren Zhou, and Junyang Lin. 2024.
\newblock \href {https://arxiv.org/abs/2409.12186} {Qwen2.5-coder technical report}.
\newblock \emph{Preprint}, arXiv:2409.12186.

\bibitem[{Os{\'e}s~Grijalba et~al.(2024)Os{\'e}s~Grijalba, Ure{\~n}a-L{\'o}pez, Mart{\'i}nez~C{\'a}mara, and Camacho-Collados}]{oses-grijalba-etal-2024-question}
Jorge Os{\'e}s~Grijalba, L.~Alfonso Ure{\~n}a-L{\'o}pez, Eugenio Mart{\'i}nez~C{\'a}mara, and Jose Camacho-Collados. 2024.
\newblock \href {https://aclanthology.org/2024.lrec-main.1179/} {Question answering over tabular data with {D}ata{B}ench: A large-scale empirical evaluation of {LLM}s}.
\newblock In \emph{Proceedings of the 2024 Joint International Conference on Computational Linguistics, Language Resources and Evaluation (LREC-COLING 2024)}, pages 13471--13488, Torino, Italia. ELRA and ICCL.

\bibitem[{Os\'es~Grijalba et~al.(2025)Os\'es~Grijalba, Ureña-L\'opez, Mart\'inez~C\'amara, and Camacho-Collados}]{osesgrijalba-etal-2025-semeval-2025}
Jorge Os\'es~Grijalba, L.~Alfonso Ureña-L\'opez, Eugenio Mart\'inez~C\'amara, and Jose Camacho-Collados. 2025.
\newblock \href {https://aclanthology.org/2025.semeval2025-1.135} {Semeval-2025 task 8: Question answering over tabular data}.
\newblock In \emph{Proceedings of the 19th International Workshop on Semantic Evaluation (SemEval-2025)}, pages 1015--1022, Vienna, Austria. Association for Computational Linguistics.

\bibitem[{Pasupat and Liang(2015)}]{pasupat2015compositionalsemanticparsingsemistructured}
Panupong Pasupat and Percy Liang. 2015.
\newblock \href {https://arxiv.org/abs/1508.00305} {Compositional semantic parsing on semi-structured tables}.
\newblock \emph{Preprint}, arXiv:1508.00305.

\bibitem[{Touvron et~al.(2023)Touvron, Lavril, Izacard, Martinet, Lachaux, Lacroix, Rozière, Goyal, Hambro, Azhar, Rodriguez, Joulin, Grave, and Lample}]{touvron2023llamaopenefficientfoundation}
Hugo Touvron, Thibaut Lavril, Gautier Izacard, Xavier Martinet, Marie-Anne Lachaux, Timothée Lacroix, Baptiste Rozière, Naman Goyal, Eric Hambro, Faisal Azhar, Aurelien Rodriguez, Armand Joulin, Edouard Grave, and Guillaume Lample. 2023.
\newblock \href {https://arxiv.org/abs/2302.13971} {Llama: Open and efficient foundation language models}.
\newblock \emph{Preprint}, arXiv:2302.13971.

\bibitem[{Vaswani et~al.(2017)Vaswani, Shazeer, Parmar, Uszkoreit, Jones, Gomez, Kaiser, and Polosukhin}]{vaswani2017attentionneed}
Ashish Vaswani, Noam Shazeer, Niki Parmar, Jakob Uszkoreit, Llion Jones, Aidan~N. Gomez, Lukasz Kaiser, and Illia Polosukhin. 2017.
\newblock \href {https://arxiv.org/abs/1706.03762} {Attention is all you need}.
\newblock \emph{Preprint}, arXiv:1706.03762.

\bibitem[{Wang et~al.(2015)Wang, Berant, and Liang}]{wang-etal-2015}
Yushi Wang, Jonathan Berant, and Percy Liang. 2015.
\newblock \href {https://doi.org/10.3115/v1/P15-1129} {Building a semantic parser overnight}.
\newblock In \emph{Proceedings of the 53rd Annual Meeting of the Association for Computational Linguistics and the 7th International Joint Conference on Natural Language Processing (Volume 1: Long Papers)}, pages 1332--1342, Beijing, China. Association for Computational Linguistics.

\bibitem[{Wang et~al.(2021)Wang, Dong, Jia, Li, Fu, Han, and Zhang}]{Wang_2021}
Zhiruo Wang, Haoyu Dong, Ran Jia, Jia Li, Zhiyi Fu, Shi Han, and Dongmei Zhang. 2021.
\newblock \href {https://doi.org/10.1145/3447548.3467434} {Tuta: Tree-based transformers for generally structured table pre-training}.
\newblock In \emph{Proceedings of the 27th ACM SIGKDD Conference on Knowledge Discovery \& Data Mining}, KDD ’21. ACM.

\bibitem[{Yang et~al.(2024)Yang, Tang, Zhao, Xiao, and Lin}]{yang-etal-2024-effective}
Bohao Yang, Chen Tang, Kun Zhao, Chenghao Xiao, and Chenghua Lin. 2024.
\newblock \href {https://aclanthology.org/2024.lrec-main.492/} {Effective distillation of table-based reasoning ability from {LLM}s}.
\newblock In \emph{Proceedings of the 2024 Joint International Conference on Computational Linguistics, Language Resources and Evaluation (LREC-COLING 2024)}, pages 5538--5550, Torino, Italia. ELRA and ICCL.

\bibitem[{Yin et~al.(2020)Yin, Neubig, tau Yih, and Riedel}]{yin2020tabertpretrainingjointunderstanding}
Pengcheng Yin, Graham Neubig, Wen tau Yih, and Sebastian Riedel. 2020.
\newblock \href {https://arxiv.org/abs/2005.08314} {Tabert: Pretraining for joint understanding of textual and tabular data}.
\newblock \emph{Preprint}, arXiv:2005.08314.

\bibitem[{Zhang et~al.(2024)Zhang, Luu, and Zhao}]{zhang-etal-2024-syntqa}
Siyue Zhang, Anh~Tuan Luu, and Chen Zhao. 2024.
\newblock \href {https://doi.org/10.18653/v1/2024.findings-emnlp.131} {{S}yn{TQA}: Synergistic table-based question answering via mixture of text-to-{SQL} and {E}2{E} {TQA}}.
\newblock In \emph{Findings of the Association for Computational Linguistics: EMNLP 2024}, pages 2352--2364, Miami, Florida, USA. Association for Computational Linguistics.

\end{thebibliography}

\clearpage
\appendix
\onecolumn

\section*{Appendix}
\label{sec:appendix}

\section{Example Schemas}
\label{sec:example-schemas}

\lstdefinestyle{sqlstyle}{
  language=SQL,
  basicstyle=\ttfamily\small,
  commentstyle=\color{gray},
  keywordstyle=\color{blue},
  breaklines=true,
  frame=single,
  captionpos=b
}

% Define a style for plain text (for Pandas schemas)
\lstdefinestyle{textstyle}{
  basicstyle=\ttfamily\small,
  breaklines=true,
  frame=single,
  captionpos=b
}

% ==============================
% Section for 067_TripAdvisor
% ==============================
\subsection{\texttt{067\_TripAdvisor}}
\label{sec:TripAdvisor-schema}

\lstset{style=textstyle}
\begin{lstlisting}
"Here are the columns for the dataset
Column Name: ratings, Data type -- object, -- Example values: {'service': 5.0, 'cleanliness': 5.0, 'overall': 5.0, 'value': 4.0, 'location': 5.0, 'sleep_qualit..., Total unique elements: 5530
Column Name: title, Data type -- category, -- Example values: ``Very nice experience for a country boy going to town'', Total unique elements: 17747
Column Name: text, Data type -- object, -- Example values: Being from a small town in Tennessee, I was very unsure of what to expect from the large city hot..., Total unique elements: 20000
Column Name: author, Data type -- object, -- Example values: {'username': 'Tucker124', 'num_reviews': 1, 'id': '39AA7B174D045F1E2BAE8A398D00BBC2', 'location':..., Total unique elements: 17995
Column Name: date_stayed, Data type -- category, -- Example values: October 2010, October 2009, September 2007, February 2012, Total unique elements: 121
Column Name: offering_id, Data type -- uint32, -- Example values: 111492, 108562, 94354, 98798, 93889, Total unique elements: 2651
Column Name: num_helpful_votes, Data type -- uint8, -- Example values: 2, 0, 1, 3, 5, Total unique elements: 40
Column Name: date, Data type -- datetime64[ns, UTC], -- Example values: 2010-10-25 00:00:00+00:00, 2009-10-14 00:00:00+00:00, 2007-10-20 00:00:00+00:00, Total unique elements: 3082
Column Name: id, Data type -- uint32, -- Example values: 84800976, 46861760, 10172355, 124329781, 69904714, Total unique elements: 20000
Column Name: via_mobile, Data type -- bool, -- Example values: False, True, Total unique elements: 2"
\end{lstlisting}

% ==============================
% Section for 069_Taxonomy
% ==============================
\subsection{\texttt{069\_Taxonomy}}
\label{sec:Taxonomy-schema}

\lstset{style=textstyle}
\begin{lstlisting}
Here are the columns for the dataset
Column Name: unique_id, Data type -- float64, -- Example values: 150.0, 151.0, 179.0, 181.0, 153.0, Total unique elements: 672
Column Name: parent, Data type -- category, -- Example values: 150, 1, 2, 37, 16, Total unique elements: 85
Column Name: name, Data type -- category, -- Example values: Attractions, Amusement and Theme Parks, Bars & Restaurants, Total unique elements: 703
Column Name: tier_1, Data type -- category, -- Example values: Attractions, Automotive, Books and Literature, Business and Finance, Total unique elements: 40
Column Name: tier_2, Data type -- category, -- Example values: Amusement and Theme Parks, Bars & Restaurants, Casinos & Gambling, Total unique elements: 347
Column Name: tier_3, Data type -- category, -- Example values: Commercial Trucks, Convertible, Coupe, Crossover, Hatchback, Total unique elements: 256
Column Name: tier_4, Data type -- category, -- Example values: Angel Investment, Bankruptcy, Business Loans, Debt Factoring & Invoice Discounting, Total unique elements: 60
Column Name: unnamed_7, Data type -- category, -- Example values: SCD, Total unique elements: 1"
\end{lstlisting}

\clearpage
% ==============================
% Section for 076_NBA
% ==============================
\subsection{\texttt{076\_NBA}}
\label{sec:NBA-schema}

\lstset{style=textstyle}
\begin{lstlisting}
Here are the columns for the dataset 
Column Name: year, Data type -- category, -- Example values: 2012-13, 2013-14, 2014-15, 2015-16, 2016-17, Total unique elements: 12
Column Name: season_type, Data type -- category, -- Example values: Regular%20Season, Playoffs, Total unique elements: 2
Column Name: player_id, Data type -- uint32, -- Example values: 201142, 977, 2544, 201935, 2546, Total unique elements: 1572
Column Name: rank, Data type -- uint16, -- Example values: 1, 2, 3, 4, 5, Total unique elements: 546
Column Name: player, Data type -- category, -- Example values: Kevin Durant, Kobe Bryant, LeBron James, James Harden, Carmelo Anthony, Total unique elements: 1568
Column Name: team_id, Data type -- uint32, -- Example values: 1610612760, 1610612747, 1610612748, 1610612745, 1610612752, Total unique elements: 30
Column Name: team, Data type -- category, -- Example values: OKC, LAL, MIA, HOU, NYK, Total unique elements: 31
Column Name: gp, Data type -- uint8, -- Example values: 81, 78, 76, 67, 82, Total unique elements: 84
Column Name: min, Data type -- uint16, -- Example values: 3119, 3013, 2877, 2985, 2482, Total unique elements: 2474
Column Name: fgm, Data type -- uint16, -- Example values: 731, 738, 765, 585, 669, Total unique elements: 697
Column Name: fga, Data type -- uint16, -- Example values: 1433, 1595, 1354, 1337, 1489, Total unique elements: 1263
Column Name: fg_pct, Data type -- float64, -- Example values: 0.51, 0.463, 0.565, 0.438, 0.449, Total unique elements: 500
Column Name: fg3m, Data type -- uint16, -- Example values: 139, 132, 103, 179, 157, Total unique elements: 274
Column Name: fg3a, Data type -- uint16, -- Example values: 334, 407, 254, 486, 414, Total unique elements: 598
Column Name: fg3_pct, Data type -- float64, -- Example values: 0.416, 0.324, 0.406, 0.368, 0.379, Total unique elements: 386
Column Name: ftm, Data type -- uint16, -- Example values: 679, 525, 403, 674, 425, Total unique elements: 447
Column Name: fta, Data type -- uint16, -- Example values: 750, 626, 535, 792, 512, Total unique elements: 541
Column Name: ft_pct, Data type -- float64, -- Example values: 0.905, 0.839, 0.753, 0.851, 0.83, Total unique elements: 552
Column Name: oreb, Data type -- uint16, -- Example values: 46, 66, 97, 62, 134, Total unique elements: 292
Column Name: dreb, Data type -- uint16, -- Example values: 594, 367, 513, 317, 326, Total unique elements: 616
Column Name: reb, Data type -- uint16, -- Example values: 640, 433, 610, 379, 460, Total unique elements: 774
Column Name: ast, Data type -- uint16, -- Example values: 374, 469, 551, 455, 171, Total unique elements: 573
Column Name: stl, Data type -- uint8, -- Example values: 116, 106, 129, 142, 52, Total unique elements: 165
Column Name: blk, Data type -- uint16, -- Example values: 105, 25, 67, 38, 32, Total unique elements: 181
Column Name: tov, Data type -- uint16, -- Example values: 280, 287, 226, 295, 175, Total unique elements: 296
Column Name: pf, Data type -- uint16, -- Example values: 143, 173, 110, 178, 205, Total unique elements: 276
Column Name: pts, Data type -- uint16, -- Example values: 2280, 2133, 2036, 2023, 1920, Total unique elements: 1539
Column Name: eff, Data type -- int16, -- Example values: 2462, 1921, 2446, 1872, 1553, Total unique elements: 1674
Column Name: ast_tov, Data type -- float64, -- Example values: 1.34, 1.63, 2.44, 1.54, 0.98, Total unique elements: 470
Column Name: stl_tov, Data type -- float64, -- Example values: 0.41, 0.37, 0.57, 0.48, 0.3, Total unique elements: 236
\end{lstlisting}

\section{Code Generation Prompts}
\label{sec:codegen}
\subsection{Pandas Code Generation without Error Handling}

\begin{tcolorbox}[title=Natural Language to Python Code with Pandas]
Generate a python code to answer this question: \texttt{\{question\}} that strictly follows the instructions below:\\

The code should return a print statement with the answer to the question.\\
The code should leave the answer be and not print anything other than the variable that holds the answer.\\
Please write a single Python code block that answers the following question and prints the result in one line at the end.\\
If the question doesn't specifically ask for it, don't use unique() or drop\_duplicates() functions.\\
\\
If it is a Yes or No question, the answer should be a boolean.\\
Do not include any explanations, comments, or additional code blocks.\\
Do not print intermediate steps just the answer.\\
Do not interact with the user.\\
Never display any sort of dataframes or tables.\\
Your output can never take more than a single line after printing and it can never be any sort of objects such as pandas or numpy objects, series etc.\\
Your output must be one of the following:\\
\\
Boolean: True/False\\
Category/String: A value\\
Number: A numerical value\\
List[category/string]: ['cat', 'dog']\\
List[number]: [1, 2, 3]\\
So the outputs have to be native python\\
\\
Given the dataset schema \{schema\}\\
\\
The following python code made for pandas for the parquet file \{dataset\_name\}.parquet reads the parquet file and running it returns the answer that is enough to answer the question \texttt{\{question\}}
\end{tcolorbox}

\subsection{Pandas Code Generation with Error Handling}
The following prompt replaces the part after the schema is given of the previous prompt.  

\begin{tcolorbox}[title=Natural Language to Python Code with Pandas - Error Correction]
The following codes generated an error when executed:
\begin{verbatim}
{code_1}/{error_1},
{code_2}/{error_2},
... % 
\end{verbatim}

Error: \{error\_msg\} Solve the error and provide the corrected code

The following python code made for pandas for the parquet file \{dataset\_name\}.parquet reads the parquet file and running it returns the answer that is enough to answer the question \texttt{\{question\}} with the error fixed
\end{tcolorbox}

\newpage
\section{Error Analysis}
\label{sec:error-analysis}

\begin{figure*}[h!]
    \centering
    \includegraphics[width=1\linewidth]{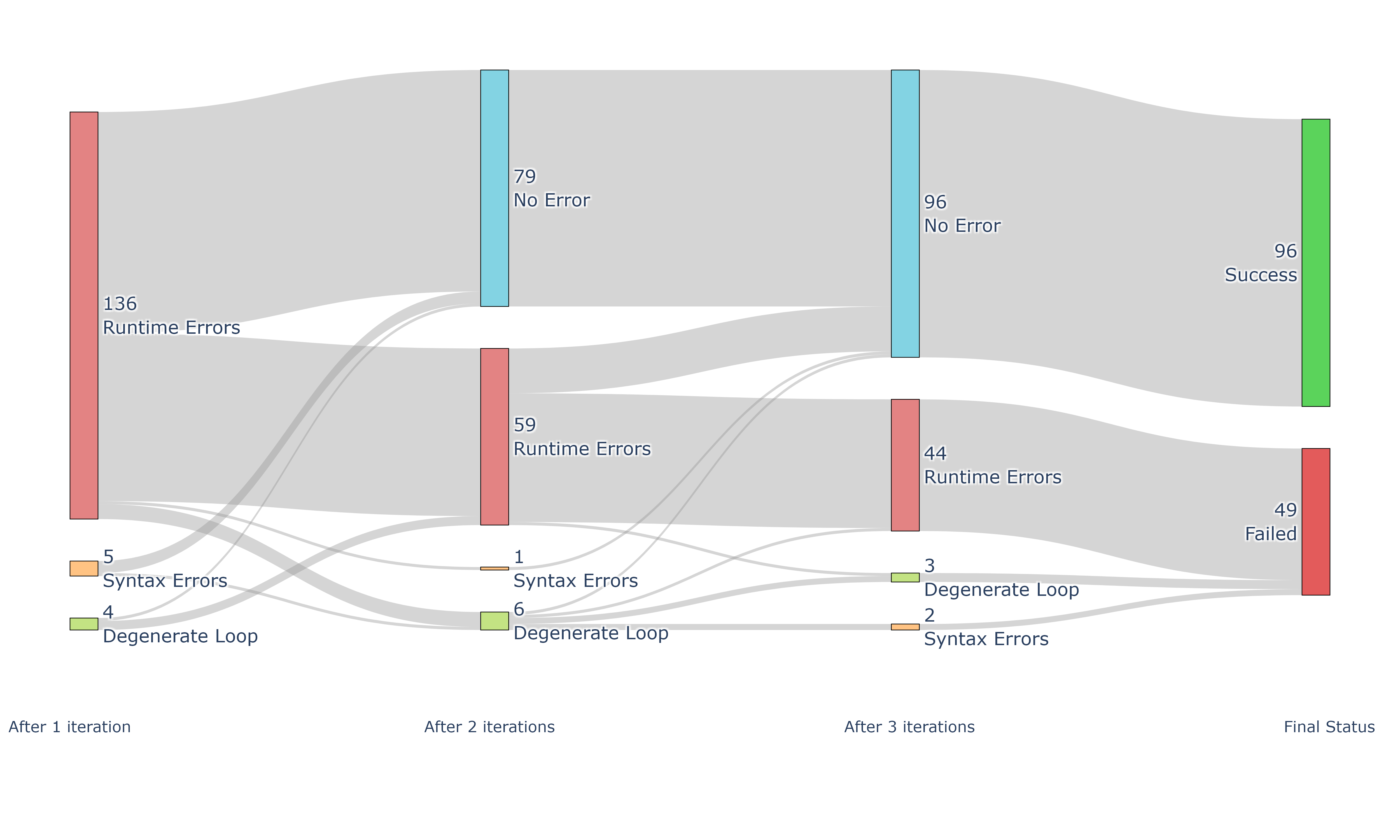}
    \caption{Error evolution and resolution across iterations (Aggregated over all models).}
    \label{fig:system-arch2}
\end{figure*}

\begin{figure*}[h!]
    \centering
    \includegraphics[width=1\linewidth]{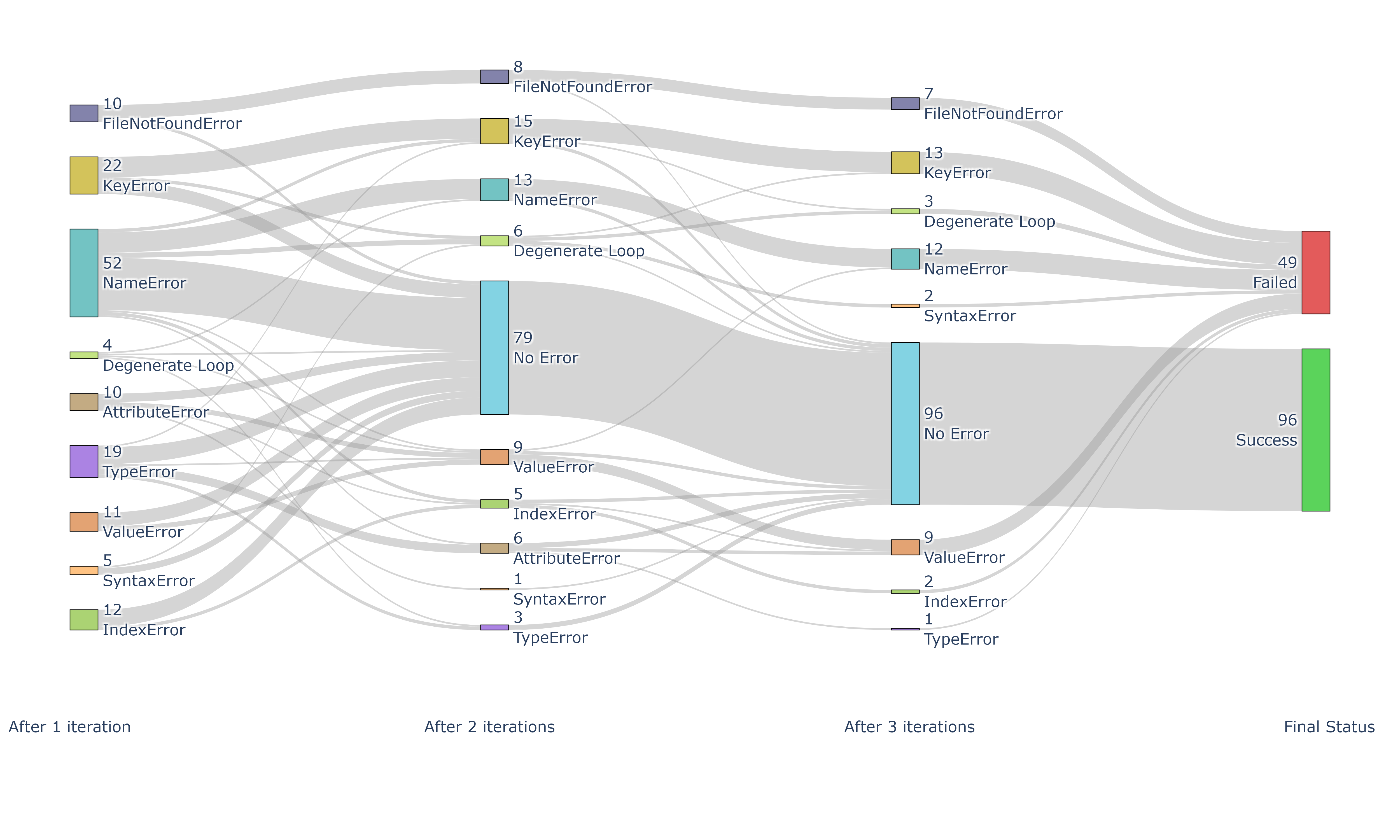}
    \caption{Fine-grained error evolution across iterations (Runtime error breakdown).}
    \label{fig:system-arch3}
\end{figure*}

\clearpage
\begin{table*}[h!]
  \centering
  \small
  \begin{tabular}{lcccccc}
    \toprule
    \textbf{Models} & \textbf{Iteration} & \textbf{Runtime} & \textbf{Degenerate Loop} & \textbf{Syntax} & \textbf{Total} \\
    \midrule
    DeepSeek-R1 (Dev)   & 1 & 9  & 0 & 0  & 9  \\
                       & 2 & 4  & 2 & 1 &  7  \\
                       & 3 & 3  & 1 & 0  & 4  \\
    \midrule
    DeepSeek-R1 (Test)   & 1 & 9  & 4 & 2 &  15 \\
                       & 2 & 6  & 1 & 0 &  7  \\
                       & 3 & 4  & 0 & 1 &  5  \\
    \midrule
    DeepSeek-V3 (Dev)    & 1 & 35 & 0 & 0 &  35 \\
                       & 2 & 8  & 3 & 0 &  11 \\
                       & 3 & 8  & 2 & 1 &  11 \\
    \midrule
    DeepSeek-V3 (Test)   & 1 & 15 & 0 & 3 &  18 \\
                       & 2 & 9  & 0 & 0 &  9  \\
                       & 3 & 5  & 0 & 0 &  5  \\
    \midrule
    Llama-3.3-70B-Instruct (Dev)      & 1 & 16 & 0 & 0 & 16 \\
                       & 2 & 5  & 0 & 0 & 5  \\
                       & 3 & 2  & 0 & 0 & 2  \\
    \midrule
    Llama-3.3-70B-Instruct (Test)     & 1 & 16 & 0 & 0 & 16 \\
                       & 2 & 10 & 0 & 0 & 10 \\
                       & 3 & 9  & 0 & 0 & 9  \\
    \midrule
    Qwen2.5-Coder-32B-Instruct (Dev)  & 1 & 11 & 0 & 0 & 11 \\
                       & 2 & 9  & 0 & 0 & 9  \\
                       & 3 & 8  & 0 & 0 & 8  \\
    \midrule
    Qwen2.5-Coder-32B-Instruct (Test) & 1 & 25 & 0 & 0 & 25 \\
                       & 2 & 8  & 0 & 0 & 8  \\
                       & 3 & 5  & 0 & 0 & 5  \\
    \bottomrule
  \end{tabular}
  \caption{Top error types and their distribution across iterations.}
  \label{tab:error-distribution}
\end{table*}

\end{document}